\documentclass[a4paper,oneside,english,oneside,reqno,twoside]{amsart}
\usepackage[T1]{fontenc}
\usepackage[latin9]{inputenc}
\usepackage{float}
\usepackage{units}
\usepackage{amsthm}
\usepackage{amstext}
\usepackage{graphicx}
\usepackage{amssymb}

\makeatletter
\numberwithin{equation}{section} 
\numberwithin{figure}{section} 
\theoremstyle{plain}

\usepackage{subfig}

\usepackage{babel}

\begin{document}

\title{RANDOM SCATTERING OF BITS BY PREDICTION}

\author{Joel Ratsaby}

\thanks{A shorter version of the paper was presented as: .J. Ratsaby. Randomness
properties of statistical prediction. \emph{The 55th meeting of the
Israel physics society (IPS 2009), Bar Ilan University}, Dec. 13,
2009.}

\address{Department of Electrical and Electronics Engineering, Ariel University
Center, Ariel 40700, ISRAEL}

\email{\texttt{ratsaby@ariel.ac.il} }
\begin{abstract}
We investigate a population of binary mistake sequences that result
from learning with parametric models of different order. We obtain
estimates of their error, algorithmic complexity and divergence from
a purely random Bernoulli sequence. We study the relationship of these
variables to the learner's information density parameter which is
defined as the ratio between the lengths of the compressed to uncompressed
files that contain the learner's decision rule. The results indicate
that good learners have a low information density$\rho$ while bad
learners have a high $\rho$. Bad learners generate mistake sequences
that are atypically complex or diverge stochastically from a purely
random Bernoulli sequence. Good learners generate typically complex
sequences with low divergence from Bernoulli sequences and they include
mistake sequences generated by the Bayes optimal predictor. Based
on the static algorithmic interference model of \cite{Ratsaby_entropy}
the learner here acts as a static structure which {}``scatters''
the bits of an input sequence (to be predicted) in proportion to its
information density $\rho$ thereby deforming its randomness characteristics.
\end{abstract}


\maketitle

\section{\label{sec:Overview}Overview}

Ratsaby \cite{Ratsaby_entropy} introduced a quantitative definition
of the information content of a general static system (e.g. a solid
or some fixed structure) and explained how it algorithmically interferes
with input excitations thereby influencing its stability. His model
is based on concepts of the theory of algorithmic information and
randomness. He modeled a system as a selection rule of a finite algorithmic
complexity which acts on an incoming sequence of random external excitations
by selecting a subsequence as output. As postulated in \cite{Ratsaby_entropy}
a simple structure is one whose information content is small. Its
selection behavior is of low complexity since it can be more concisely
described. Consequently it is less able to deform properties of randomness
of the input sequence. And vice versa, if the system is sufficiently
complex it can significantly deform the randomness at the input. Following
\cite{Ratsaby_entropy} there have been recent theoretical and empirical
results that validate his model for specific problem domains. The
first empirical proof of his model appeared in \cite{Ratsaby_AIPR09,RatsabyChaskJSSC,RatsabyChaskCRM09}
where it was shown that this inverse relationship between system complexity
and randomness exists also in a real physical system. The particular
system investigated consisted of a one-dimensional vibrating solid-beam
to which a random sequence of external input forces is applied. In
\cite{rat0903,ratsabyICCC09} the problem of learning to predict binary
sequences was shown to be an exemplar of this paradigm. The complexity
of a learner's decision rule is proportional to the amount that the
subsequence selected by the learner (via his mistakes) deviates from
a truly random sequence. A first empirical investigation of this learning
problem appeared in \cite{Ratsaby_BRAIN10,Ratsabyuics09,Ratsaby_Acta}
where a new measure of system complexity called the sysRatio was introduced
and shown to be a proper measure of a learner's decision complexity. 

The current paper digs further along this line and provides not only
further empirical analysis and justification of the model of \cite{Ratsaby_entropy}
applied to the problem of learning but also gives new interpretations
of standard learning phenomena such as model data underfitting or
overfitting. It is shown that these phenomena can be interpreted as
certain types of deformations of randomness of the binary mistake
sequences. These deformations are measured in the $\Delta\ell$-plane
($\Delta$ stands for divergence and $\ell$ for estimated Kolmogorov
complexity). We conclude that the prediction rule obtained by learning
is analogous to a physical static object that scatters a random beam
of particles. We call this phenomena \emph{bit-scattering} (we discuss
this phenomenon later at the end of section \ref{sec:Experimentl-setup}).
The current paper is a further justification that the static algorithmic
interference model defined in \cite{Ratsaby_entropy} applies to the
problem of learning to predict. Before proceeding to give an introduction
to the main concepts let us state the problem that we consider in
the paper. 

\emph{Statement of the problem}: Given a random source that generates
two binary sequences, $x^{(m)}$ and $x^{(n)}$ of length $m$ and
$n$, respectively, according to a finite Markov chain of unknown
order $k^{*}$ with an unknown probability transition matrix. A learner
uses $x^{(m)}$ to estimate the probability parameters of a Markov
model of order $k$. Once the model is learnt, the learner makes a
prediction for every bit in $x^{(n)}$. Denote by $y^{(n)}$ the binary
sequence corresponding to these predictions. Denote by $\xi^{(n)}$
the error sequence that corresponds to the learner's predictions where
the $i^{th}$ bit $\xi_{i}=1$ if the prediction differs from the
true value, i.e., $y_{i}\neq x_{i}$ and $\xi_{i}=0$ otherwise. Denote
by $\xi_{0}^{(n)}$ the subsequence of $\xi^{(n)}$ corresponding
to those bits of $y^{(n)}$ that are $0$. In this paper we study
different characteristics of the error sequence $\xi_{0}^{(n)}$ and
how they depend on the two main learner's parameters, the training
sequence length $m$ and the model order $k$. We focus on two main
characteristics, the algorithmic complexity of the error sequence
and the statistical deviation between the frequency of $1$s and the
probability of seeing a $1$ in the sequence. We determine their interrelationship
and how the probability of a prediction error depends on them. 

The remainder of the paper is organized as follows: in section \ref{sec1}
we introduce the basic concepts of algorithmic complexity and related
properties of randomness. In section \ref{sec:Selection-rule} we
review the concept of a selection rule, in section \ref{sec:Relationship-to-information}
we state a relationship between the complexity of a finite random
binary sequence and its entropy. Section \ref{sec:Experimentl-setup}
describes the experimental setup used for the analysis followed by
section \ref{sec:Results} which describes the results. 

Before continuing, we should clarify at this point that our use of
the words 'chaoticity' or 'chaotic' is different from chaos theory.
By a chaotic binary sequence we do not necessarily mean that it is
generated by some dynamical system that is highly sensitive to initial
conditions but that it is highly disordered, or in other words, has
a high algorithmic complexity.

\section{\label{sec1}Introduction}

Algorithmic randomness (see \cite{Calude2002,Hutter2004,LaurentBienvenu07})
is a notion of randomness of an individual element (object) of a sample
space. It reflects how chaotic, or how complicated it is to describe
the object. Classical probability theory assigns probabilities to
sets of outcomes of random trials in an experiment. For instance,
consider an experiment with $n$ randomly and independently drawn
binary numbers $X_{i}$, $i=1,\ldots n$, where $X_{i}=1$ with probability
$\nicefrac{1}{2}$. Then any outcome such as $X=\left(0,0,\ldots,0\right)$
has the same probability $2^{-n}$. However, from an algorithmic perspective,
it is clear that the string $X=\left(0,0,\ldots0\right)$ is not random
compared to some other possible string with a more complicated pattern
of zeros and ones. Algorithmic randomness of finite objects (binary
sequences) aims to explain the intuitive idea that a sequence, whether
finite or infinite, should be measured as being more unpredictable
if it possess fewer regularities (patterns). There is no formal definition
of randomness but there are three main properties that a random binary
string of length $n$ must intuitively satisfy \cite{UspenskilETAL1990}.
The first property is the so-called \emph{stochasticity} or frequency
stability of the sequence which means that any binary word of length
$k\leq n$ must have the same frequency limit (equal to $2^{-k}$).
This is basically the notion of normality that Borel introduced and
is related to the degree of unpredictability of the sequence. The
second property is \emph{chaoticity} or disorderliness of the sequence.
A sequence is less chaotic (less complex) if it has a short description,
i.e., if the minimal length of a program that generates the sequence
is short. The third property is \emph{typicalness}. A random sequence
is a typical representative of the class $\Omega$ of all binary sequences.
It has no specific features distinguishing it from the rest of the
population. An infinite binary sequence is typical if each small subset
$E$ of $\Omega$ does not contain it (the correct definition of a
'small' set was given by Martin L$\ddot{o}$f \cite{MartinLof66}).

Algorithmic randomness was first considered by von Mises in 1919 who
defined an infinite binary sequence $\alpha$ of zeros and ones as
random if it is unbiased, i.e. if the frequency of zeros goes to $\nicefrac{1}{2}$,
and every subsequence of $\alpha$ that we can extract using an admissible
selection rule (see definition below) is also not biased. Kolmogorov
and Loveland \cite{Loveland66,Kolmogorov98} proposed a more permissive
definition of an admissible \emph{selection rule} as any (partial)
computable process which, having read any $n$ bits of an infinite
binary sequence $\alpha$, picks a bit that has not been read yet,
decides whether it should be selected or not, and then reads its value.
When subsequences selected by such a selection rule pass the unbiasedness
test they are called Kolmogorov-Loveland stochastic (KL-stochastic
for short). Martin L$\ddot{o}$f \cite{MartinLof66} introduced a
notion of randomness which is now considered by many as the most satisfactory
notion of algorithmic randomness. His definition says precisely which
infinite binary sequences are random and which are not. The definition
is probabilistically convincing in that it requires each random sequence
to pass every algorithmically implementable statistical test of randomness. 

In this paper we are concerned with random sequences that arise from
the process of learning and prediction, or more specifically, from
the prediction mistakes made by a learner. Let $X^{(n)}=X_{1},\ldots,X_{n}$
be a sequence of binary random variables drawn according to some unknown
joint probability distribution $\mathbb{P}\left(X^{(n)}\right)$.
Consider the problem of learning to predict the next bit in a binary
sequence drawn according to $\mathbb{P}$. For training, the learner
is given a finite sequence $x^{(m)}$ of bits $x_{t}\in\left\{ 0,1\right\} ,$
$1\leq t\leq m$, drawn according to $\mathbb{P}$ and estimates a
model $\mathcal{M}$ that can be used to predict the next bit of a
partially observed sequence. After training, the learner is tested
on another sequence $x^{(n)}$ drawn according to the same unknown
distribution $\mathbb{P}$. Using $\mathcal{M}$ he produces the bit
$y_{t}$ as a prediction for $x_{t}$ , $1\leq t\leq n$. Denote by
$\xi^{(n)}$ the corresponding binary sequence of mistakes where $\xi_{t}=1$
if $y_{t}\neq x_{t}$ and is $0$ otherwise. Denote by $\xi_{0}^{(n)}$
the subsequence of $\xi^{(n)}$ that corresponds to the times $t$
where the learner predicted $y_{t}=0$. Note that $\xi_{0}^{(n)}$
is also a subsequence of $x^{(n)}$ so we can view the process of
predicting as a process of selecting a subsequence of the input $x^{(n)}$.

It is clear that the subsequence $\xi_{0}^{(n)}$ of mistakes should
be random since the test sequence $x^{(n)}$ is random. It is reasonable
to expect that the learner may implicitly vary some of the randomness
characteristics of the subsequence of bits that he selects thereby
cause $\xi_{0}^{(n)}$ to be less random than $x^{(n)}$. In this
sense, we may say that the learner 'deforms' the randomness of the
input $x^{(n)}$ producing a less random subsequence of $x^{(n)}$.
Or perhaps the learner being of a finite complexity is limited in
his ability to 'deform' randomness of $x^{(n)}$. Essentially we ask
what 'interference' does a learner have on the randomness of a test
sequence. It appears essential that we look not only on the randomness
of the object itself (the test sequence $x^{(n)}$) but also at the
interfering entity---the learner, specifically, its algorithmic component
that is used for prediction.

\section{\label{sec:Selection-rule}Selection rule}

Let us formally define a selection rule. This is a principal concept
used as part of tests of randomness of sequences (mentioned above).
Let $\left\{ 0,1\right\} ^{*}$ be the space of all finite binary
sequences and denote by $\left\{ 0,1\right\} ^{n}$ the set of all
finite binary sequences of length $n$. An admissible \emph{selection
rule} $R$ is defined \cite{Kolmogorov98,Vyugin99} based on three
partial recursive functions $f,g$ and $h$ on $\left\{ 0,1\right\} ^{*}$.
Let $x^{(n)}=x_{1},\ldots,x_{n}$. The process of selection is recursive.
It begins with an empty sequence $\emptyset$. The function $f$ is
responsible for selecting possible candidate bits of $x^{(n)}$ as
elements of the subsequence to be formed. The function $g$ examines
the value of these bits and decides whether to include them in the
subsequence. Thus $f$ does so according to the following definition:
$f(\emptyset)=i_{1}$, and if at the current time $k$ a subsequence
has already been selected which consists of elements $x_{i_{1}},\ldots,x_{i_{k}}$
then $f$ computes the index of the next element to be examined according
to element $f(x_{i_{1}},\ldots,x_{i_{k}})=i$ where $i\not\in\{i_{1},\ldots,i_{k}\}$,
i.e., the next element to be examined must not be one which has already
been selected (notice that maybe $i<i_{j}$, $1\leq j\leq k$, i.e.,
the selection rule can go backwards on $x$). Next, the two-valued
function $g$ selects this element $x_{i}$ to be the next element
of the constructed subsequence of $x$ if and only if $g(x_{i_{1}},\ldots,x_{i_{k}})=1$.
The role of the two-valued function $h$ is to decide when this process
must be terminated. This subsequence selection process terminates
if $h(x_{i_{1}},\ldots,x_{i_{k}})=1$ or $f(x_{i_{1}},\ldots,x_{i_{k}})>n$.
Let $R(x^{(n)})$ denote the selected subsequence. By $K(R|n)$ we
mean the length of the shortest program computing the values of $f$,
$g$ and $h$ given $n$.

From the above discussion, we know that there are two principal measures
related to the information content in a finite sequence $x^{(n)}$,
stochasticity (unpredictability) and chaoticity (complexity). An infinitely
long binary sequence is regarded random if it satisfies the principle
of stability of the frequency of $1$s for any of its subsequences
that are obtained by an admissible selection rule \cite{Kolmogorov98}.
Kolmogorov showed that the stochasticity of a finite binary sequence
$x$ may be precisely expressed by the deviation of the frequency
of ones from some $0<p<1$, for any subsequence of $x^{(n)}$ selected
by an admissible selection rule $R$ of finite complexity $K(R|n)$
where for an object $x$ given another object $y$ he defined in \cite{Kolmogorov65}
the complexity of $x$ as

\begin{equation}
K(x|y)=\min\{l(\pi):\phi(\pi,y)=x\}\label{eq:K}\end{equation}
where $l(\pi)$ is the length of the sequence $\pi$, $\phi$ is a
universal partial recursive function which acts as a description method,
i.e., when provided with input $(\pi,y)$ it gives a specification
for $x$ (for an introduction see section 2 of \cite{RatsabyChaskJSSC}).
The chaoticity of $x^{(n)}$ is large if its complexity is close to
its length $n$. The classical work of \cite{Asarin87,Asarin88,Kolmogorov98,Vyugin99}
relates chaoticity to stochasticity. In \cite{Asarin87,Asarin88}
it is shown that chaoticity implies stochasticity. For a binary sequence
$s$, let us denote by $\|s\|$ the number of $1$s in $s$, then
this can be seen from the following relationship (with $p=\nicefrac{1}{2}$):

\begin{eqnarray}
\left|\frac{\|R(x^{(n)})\|}{l(R(x^{(n)}))}-\frac{1}{2}\right| & \leq & c\sqrt{\frac{n-K(x^{(n)}|n)+K(R|n)+2\log K(R|n)}{l(R(x^{(n)}))}}\label{eq:Ineq}\end{eqnarray}
where $l(R(x^{(n)}))$ is the length of the subsequence selected by
$R$ and $c>0$ is some absolute constant. Apparently as the chaoticity
of $x^{(n)}$ grows the stochasticity of the selected subsequence
$R(x^{(n)})$ grows (the bias from $\nicefrac{1}{2}$ decreases).
Also, and more relevant to the context of this paper, the information
content of the selection rule namely $K(R|n)$ has a direct effect
on this relationship: the lower $K(R|n)$ the stronger the stability
(smaller deviation of the frequency of $1$s from $\nicefrac{1}{2}$).
In \cite{Vereshchagin} the other direction which shows that stochasticity
implies chaoticity is proved. 

It was recently shown in \cite{rat0903,ratsabyICCC09} that the level
of randomness of the subsequence $\xi_{0}^{(n)}$ of $\xi^{(n)}$
which corresponds to the occurrences of mistakes in predicting $0$s
decreases relative to an increase in the complexity of the learner.
The approach taken there is to represent the learner's decision as
a selection rule that selects $\xi_{0}^{(n)}$ from $\xi^{(n)}$.
The rule's complexity is defined based on a combinatorial quantity
rather than Kolmogorov complexity but still yields a relationship
of the form of (\ref{eq:Ineq}). This relationship shows that the
possibility of deviation of the frequency of $1$s in $\xi_{0}^{(n)}$
from the probability $p_{0}$ of seeing a $1$ in $\xi_{0}^{(n)}$
grows as the complexity of the class of possible decisions grows. 

The current paper investigates this experimentally. We consider a
learner's prediction (or decision) rule which we term as \emph{system}
and study its influence on a random binary test sequence on which
prediction decisions are made. The system is based on the maximum
\emph{a posteriori} probability decision where probabilities are defined
by a statistical parametric model which is estimated from data. The
learner of this model is a computer program that trains from a given
random data sequence and then produces a decision rule by which it
is able to predict (or decide) the value of the next bit in future
(yet unseen) random binary sequences. As in \cite{rat0903,ratsabyICCC09}
we focus on Markov source and a Markov learner whose orders may differ.

\section{\label{sec:Relationship-to-information}Relationship to information
theory}

We now describe the connection between the concepts of entropy (Shannon
entropy) and algorithmic complexity. Entropy is a measure of unpredictability
of a random variable. Intuitively, we expect that the more unpredictable
a sequence of random variables the higher its algorithmic (Kolmogorov)
complexity. This is formally expressed as Theorem 14.3.1 in \cite{CoverThomas2006}
which we now state: denote by $H(X_{1})$ the entropy of a random
variable $X_{1}$ and consider a sequence of random variables $\left\{ X_{i}\right\} $
drawn i.i.d. according to the probability mass function $f(x)$, $x\in\mathcal{X}$,
where $\mathcal{X}$ is a finite alphabet. Let $f(x^{(n)})=\prod_{i=1}^{n}f(x_{i})$.
Then there exists a constant $c$ such that \[
H(X_{1})\leq\frac{1}{n}\sum_{x^{(n)}}f(x^{(n)})K(x^{(n)}|n)\leq H(X_{1})+\frac{\left(\left|\mathcal{X}\right|-1\right)\log n}{n}+\frac{c}{n}\]
 for all $n$. Consequently, the expected value $E\frac{1}{n}K(X^{(n)}|n)\rightarrow H(X_{1})$
with increasing $n$. This means that the expected value of the Kolmogorov
complexity of the sequence converges to the Shannon entropy of the
sequence with increasing $n$. 

A more relevant estimate for our work here concerns the Kolmogorov
complexity of a specific sequence (not the expected value over all
sequences). In the case of a Bernoulli random sequence $\left\{ X_{i}\right\} _{i=1}^{n}$
with probability $p=P(X_{i}=1)$ its complexity relates to the binary
entropy $H(X_{1})$ of any of the i.i.d. random variables of the sequence.
It is based on the following statement which holds even more generally
for any binary sequence of length $n$ (Theorem 14.2.5 of \cite{CoverThomas2006}):
Let $x^{(n)}=x_{1},x_{2},\ldots,x_{n}$ be a binary string then the
Kolmogorov complexity of $x^{(n)}$ is bounded as 

\begin{eqnarray}
K(x^{(n)}|n) & \leq & nH(p)+\frac{1}{2}\log_{2}n+c\label{eq:entropyest}\end{eqnarray}
where $p=\frac{1}{n}\sum_{i=1}^{n}x_{i}$, $H(p)=-p\log_{2}p-(1-p)\log_{2}(1-p)$
is the entropy of a binary random variable with probability $p$ and
$c$ is some finite positive constant independent of $n$ and of the
sequence $x^{(n)}$. In particular, we may compute this bound for
the random mistake sequences $\xi_{0}^{(n)}$ that we are interested
in. In section \ref{sec:Results} we use this as a comparison with
the empirical estimated algorithmic complexity which is obtained by
compression. We proceed to describe the setup.

\section{\label{sec:Experimentl-setup}Experimentl setup}

The learning problem consists of predicting the next bit in a given
sequence generated by a Markov chain (model) $\mathcal{M}^{*}$ of
order $k^{*}$. There are $2^{k^{*}}$ states in the model each represented
by a word of $k^{*}$ bits. During a learning problem, the source's
model is fixed. A learner, unaware of the source's model, has a Markov
model of order $k$. We denote by $p(1|i)$ the probability of transiting
from state $i$ whose binary $k$-word is $b_{i}=[b_{i}(1),\ldots,b_{i}(k)]$
to the state whose word is $[b_{i}(2),\ldots,b_{i}(k),1]$. Given
a random sequence of length $m$ generated by the source the learner
estimates its own model's parameters $p(1|i)$ by $\hat{p}(1|i)$,
$1\leq i\leq2^{k}$, which is the frequency of the event {}``$b_{i}$
is followed by a $1$'' in the training sequence. We denote by $\hat{\mathcal{M}}$
the learnt model with parameters $\hat{p}(1|i)$, $1\leq i\leq2^{k}$.
We denote by $p^{*}(1|i)$ the transition probability from state $i$
of the source model, $1\leq i\leq2^{k}$.

A simulation run is characterized by the parameters, $k$ and $m$.
It consists of a training and testing phases. In the training phase
we show the learner a binary sequence of length $m$ and he estimates
the transition probabilities. In the testing phase we show the learner
another random sequence (generated by the same source) of length $n$
and test the learner's predictions on it. For each bit in the test
sequence we record whether the learner has made a mistake. When a
mistake occurs we indicate this by a $1$ and when there is no mistake
we write a $0$. The resulting sequence of length $n$ is the generalization
mistake sequence $\xi^{(n)}$. We denote by $\xi_{0}^{(n)}$ the binary
subsequence of $\xi^{(n)}$ that corresponds to the mistakes that
occurred only when the learner predicted a $0$. Its length is denoted
by $n_{0}$. We denote by $p_{0}$ the probability of mistake when
predicting a $0$, i.e., $p_{0}$ is the probability of seeing a $1$
in the subsequence $\xi_{0}^{(n)}$.

For a fixed $k$ denote by $N_{k,m}$ the number of runs with a learner
of order $k$ and training sample of size $m$. The experimental setup
consists of $N_{k,m}=10$ runs with $1\leq k\leq10$, $m\in\left\{ 100,200,\ldots,10000\right\} $
with a total of $100\cdot10\cdot N_{k,m}=10000$ runs. The testing
sequence is of length $n=1000$. Each run results in a file called
\emph{system} which contains a binary vector $d$ whose $i^{th}$
bit represents the maximum \emph{a posteriori} decision made at state
$i$ of the learner's model, i.e., \begin{equation}
d_{i}=\left\{ \begin{array}{cc}
1 & \text{if \, }\hat{p}(1|i)>\nicefrac{1}{2}\\
0 & otherwise\end{array}\right.\label{eq:zi}\end{equation}
for $1\leq i\leq2^{k}$. Let us denote by $\alpha_{i}=P(\hat{p}(1|i)>\nicefrac{1}{2})$,
thus $d_{i}$ are Bernoulli random variables with parameters $\alpha_{i}$,
$1\leq i\leq2^{k}$. The learner's system is comprised of the decision
at every possible state.

Another file generated is the \emph{errorT0} which contains the mistake
subsequence $\xi_{0}^{(n)}$. At the end of each run we measure the
lengths of the \emph{system} file and its compressed length where
compression is obtained either via the Gzip algorithm (a variant of
\cite{LZ77}) or the PPM algorithm \cite{PPM} and compute the \emph{sysRatio}
(denoted as $\rho)$ which is the ratio of the compressed to uncompressed
length of the system file. Note that $\rho$ is a measure of information
density since it captures the number of bits of useful information
(useful for describing the system) per bit of representation (in the
uncompressed file). 

We do similarly for the mistake-subsequence $\xi_{0}^{(n)}$ obtaining
the length $\ell_{0}$ of the compressed file that contains $\xi_{0}^{(n)}$
(henceforth referred to as the estimated algorithmic complexity of
$\xi_{0}^{(n)}$ since it is an approximation of the Kolmogorov complexity
of $\xi_{0}^{(n)}$, see \cite{RatsabyChaskJSSC}). We measure the
KL-divergence $\Delta_{0}$ between the probability distribution $P(w|\hat{p})$
of binary words $w$ of length $4$ and the empirical probability
distribution $\hat{P}_{m}(w)$ as measured from the mistake subsequence
$\xi_{0}^{(n)}$. Note, $P(w|\hat{p})$ is defined according to the
Bernoulli model with parameter $\hat{p}_{0}$, that is, $P(w|\hat{p}_{0})=\hat{p}_{0}^{i}(1-\hat{p}_{0})^{4-i}$
for a word $w$ with $i$ ones, where $\hat{p}_{0}$ is the frequency
of ones in the subsequence $\xi_{0}^{(n)}$. The distribution $\hat{P}_{m}(w)$
equals the frequency of a word $w$ in $\xi_{0}^{(n)}$. Hence $\Delta_{0}$
reflects by how much $\xi_{0}^{(n)}$ deviates from being random according
to a Bernoulli sequence with parameter $p_{0}$ (the mistake probability
when predicting a $0$).

\section{\label{sec:Results}Results}

We are interested in determining the relationship between the estimated
algorithmic complexity $\ell_{0}$ of $\xi_{0}^{(n)}$, its divergence
$\Delta_{0}$ and the learning performance. As the learning performance
we look at the generalization error of type $0$ that is the error
for $0$-predictions. We choose four different levels of learning
problems, controlled by the order of the source model $k^{*}=3$,
$4$, $5$, $6$. For each problem we choose for the source model
a transition matrix of probabilities $p^{*}(1|i)=1-p$, $p^{*}(0|i)=p$,
where for some of the states $i$ we set $p=0.3$ and for others $p=0.7$,
$1\leq i\leq2^{k^{*}}$. Thus the Bayes optimal error is $0.3$. To
ensure that the problem is sufficiently challenging we set the first
half of the states (those ranging from the $k^{*}$-dimensional vector
$00\ldots0$ to $011\ldots1$) to have $p=0.3$ and the second half
($10\ldots0$ to $11\ldots1$) to have $p=0.7$. This ensures that
a Markov model of order $k<k^{*}$ cannot approximate the true transition
probabilities well. That is, the infinite-sample limit estimate based
on a Markov model of order $k$ which is smaller than $k^{*}$ will
still be $\hat{p}(1|i)=0.5$, $1\leq i\leq2^{k}$. But for a Markov
model of order $k\geq k^{*}$ the infinite-sample size estimates will
converge to the true values of $p$ or $1-p$.

\subsection{\label{sub:Learning-curves}Learning curves}

\begin{figure}
\begin{centering}
\includegraphics[clip,scale=0.6] {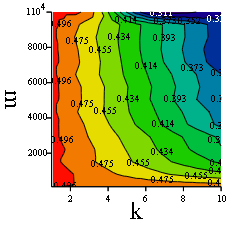}
\par\end{centering}

\caption{\label{fig:generalization-error-versus}Generalization error with
respect to $k$ and $m$ for $k^{*}=5$ }

\end{figure}

Before we start to investigate the three relationships stated above
we perform a sanity check to see how the prediction generalization
error (for any of the two prediction types, not just when predicting
a zero) varies with respect to the model complexity $k$ and training
length $m$. This is the so-called 'learning curves' in the areas
of statistical pattern recognition and learning theory \cite{AB99}.
Figure \ref{fig:generalization-error-versus} displays the contours
of the error surface as a function of $k$ and $m$ for a learning
problem with $k^{*}=5$ (the Bayes error is $0.3$). As can be seen,
when $k<k^{*}$ the error remains very high, close to $0.5$, regardless
of the training sample size $m$ (this is the leftmost contour colored
in red). For $k>k^{*}$ the prediction error gets closer to the Bayes
$0.3$ value (outermost contour colored in dark blue) with increasing
$m$. The shape of the contours indicate the tradeoff between approximation
and estimation errors whose sum is the prediction error (standard
results from learning theory, see for instance \cite{RatsabyMM96,AB99,BartlettBoucheronLugosi02}).
The larger that $k$ becomes the lower the approximation error. The
larger that $m$ becomes the smaller the estimation error. 

We now proceed to describe the main result which concerns  the relationship
between the learner's performance and the mistake sequence complexity.

\subsection{\label{sub:sysRatio--versus}sysRatio $\rho$ versus $k$}

\begin{figure}
\begin{raggedright}
\subfloat[Gzip-based compressor]{\noindent \begin{raggedright}
\includegraphics[clip,scale=0.7]{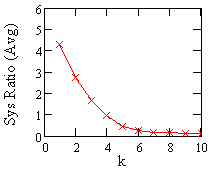}
\par\end{raggedright}

}\subfloat[PPM-based compressor]{\noindent \begin{raggedright}
\includegraphics[clip,scale=0.7]{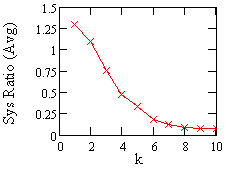}
\par\end{raggedright}

}
\par\end{raggedright}

\caption{\label{fig:SysRatio--versus K}The average sysRatio $\rho$ versus
model order $k$}

\end{figure}

First we look at the relationship between the sysRatio $\rho$ and
$k$. Figure \ref{fig:SysRatio--versus K} shows the average of the
sysRatio $\rho$ as a function of $k$ where in Figure \ref{fig:SysRatio--versus K}(A)
we used Gzip as the compressor that estimates the Kolmogorov complexity
and in Figure \ref{fig:SysRatio--versus K} (B) we used the PPM algorithm
as compressor. Note that the PPM compressor obtains $\rho$ values
that are smaller than the Gzip compressor which means that the compressed
lengths of the corresponding system files is smaller when using PPM.
We believe that this is due to additional cost incurred by Gzip in
the form of data structures that are appended to the compressed data.
This is more noticeable when the file to be compressed is small (for
instance, in the plot we see that the the sysRatio only goes below
unity at around $k=4$ which is when the uncompressed file length
goes above $16$). The PPM compressor thus approximates the algorithmic
(Kolmogorov) complexity better than Gzip when the uncompressed files
are relatively small. In the remainder of the paper we decided to
keep the plots with respect to both types of compressors in order
to show that the results of our analysis do not significantly vary
as one changes from one type of compressor to another (in some places
we put only the Gzip-based results since the differences were insignificant).

Looking at the plots of Figure \ref{fig:SysRatio--versus K} it is
clear that the average sysRatio decreases as the learner's model order
$k$ increases. For the PPM compressor, we see a critical point at
the vicinity of $k^{*}$ where the convexity of the graph changes
from concave down to concave up possibly indicating an inflection
point (this holds for learning problems with other values of $k^{*}$,
for instance in Appendix \ref{sec:apA} we show this for $k^{*}=3$
and $k^{*}=7$). To explain this, first note that the uncompressed
length of the system is always $c\cdot2^{k}$ for some constant $c>0$
since the vector $d$ is of length $2^{k}$ (see section \ref{sec:Experimentl-setup}).
The length of the compressed system file also grows, but at a slower
rate with respect to $k$ and this gives rise to the decrease in $\rho$
with respect to $k$. We can explain why the rate of the compressed
system file grows more slowly as follows: for values of $k<k^{*}$
the learner's model is incapable (by design of the learning problem)
of estimating the Bayes optimal prediction and the probability of
the events {}``$b_{i}$ is followed by a $1$'' is $p(1|i)=\nicefrac{1}{2}$
, $1\leq i\leq2^{k}$. Thus the average value $\hat{p}(1|i)$ of the
indicators of such events is a Binomial random variable with a distribution
symmetric at $\nicefrac{1}{2}$ and hence from (\ref{eq:zi}) the
probability $\alpha_{i}$ that $\hat{p}(1|i)>\nicefrac{1}{2}$ equals
$\nicefrac{1}{2}$. The components of the random vector $d$ are independent
Bernoulli random variables with parameter $\alpha_{i}$ when conditioned
on the sample size vector $v$ (this is the vector whose components
$v_{i}$ are the number of times that $b_{i}$ appeared in the training
sequence, see \cite{rat0903} for details). Since in this case $\alpha_{i}=\nicefrac{1}{2}$
then each component has a maximum entropy $H(d_{i})=-\alpha_{i}\log\alpha_{i}-(1-\alpha_{i})\log(1-\alpha_{i})=\log2=1$
and hence the expected value of the entropy of the vector $d$ (with
respect to the random sample size vector $v$) is maximal and equals
$E_{v}H(d|v)=E_{v}\sum_{i=1}^{2^{k}}H(d_{i}|v_{i})=E_{v}2^{k}=2^{k}.$
Hence the expected compressed length of the system file (which contains
the vector $d$) is large as the expected description length of any
random variable is at least as large as its entropy. 

As $k$ increases beyond $k^{*}$ the model becomes more capable of
estimating the true transition probabilities (recall, these are either
$0.3$ or $0.7$) and the probability $p(1|i)$ of the events {}``$b_{i}$
is followed by a $1$'' get farther away from $\nicefrac{1}{2}$
in the direction of $0.3$ or $0.7$, depending on the particular
state $i$, $1\leq i\leq2^{k}$. Thus the average value $\hat{p}(1|i)$
of the indicators of such events is a Binomial random variable with
an asymmetric distribution with a mean $p(1|i$). Hence from (\ref{eq:zi})
the probability $\alpha_{i}$ that $\hat{p}(1|i)>\nicefrac{1}{2}$
gets either very close to $0$ or $1$ as the training size $m$ increases.
Thus the components of the random vector $d$ tend to be closer to
deterministic. They are still random since the training sequence length
is not increasing with $k$ and the variance of the estimates $\hat{p}(1|i)$
does not converge to zero. Therefore for each of the $2^{k}$ components
of the vector $d$ the entropy is smaller than when $k<k^{*}$. However
as there are exponentially many components $d_{i}$, on the whole,
the entropy of $d$ (and hence the expected compressed length of the
system file) still increases but at a lower rate than when $k<k^{*}$. 

\begin{figure}
\begin{centering}
\includegraphics[clip,scale=0.6]{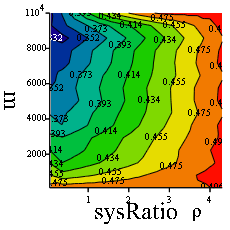}
\par\end{centering}

\caption{\label{fig:Learning-curves-with}Generalization error with respect
to $\rho$ and $m$ (for $k^{*}=5$)}

\end{figure}

We can now alternatively look at the learning curves (section \ref{sub:Learning-curves})
based on the sysRatio (instead of $k$). This is shown in Figure \ref{fig:Learning-curves-with}.
Clearly, good learners are those with low value of sysRatio $\rho$
(left uppermost region which is colored dark blue) while bad learners
are those with a high sysRatio $\rho$, displayed as the rightmost
contour which spans from lowest to highest $m$ values.

We proceed now to discuss the characteristics of the mistake subsequence
$\xi_{0}^{(n)}$. First, in section \ref{sub:Estimated-algorithmic-complexity}
we study how its estimated algorithmic complexity $\ell_{0}$ and
divergence $\Delta_{0}$ depend on the learner's decision characteristics,
or formally, the sysRatio $\rho$. In section \ref{sub:Estimated-algorithmic-complexity_DIF}
we fix the learner's model order $k$ and study how $\ell_{0}$ depends
on $\Delta_{0}$. Finally in sections \ref{sub:The-error-surface}
and \ref{sub:The-sysRatio-surface} we study the $p_{0}$ and $\rho$
surfaces over the $\Delta\ell$-plane.

\subsection{\label{sub:Estimated-algorithmic-complexity}Estimated algorithmic
complexity $\ell_{0}$ and divergence $\Delta_{0}$ versus sysRatio
$\rho$}

\begin{figure}
\subfloat[Gzip as compressor]{\begin{raggedright}
\includegraphics[clip,scale=0.6]{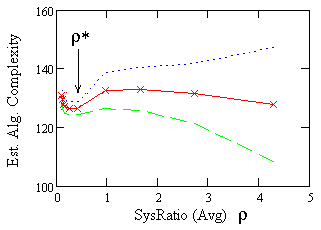}
\par\end{raggedright}

}\subfloat[PPM as compressor]{\includegraphics[clip,scale=0.6]{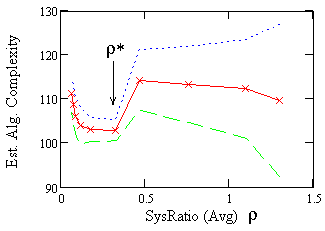}

}

\caption{\label{fig:Algorithmic-complexity-}Estimated algorithmic complexity
$\ell_{0}$ of the mistake subsequence $\xi_{0}^{(n)}$ versus the
sysRatio (average)}

\end{figure}

Note, in the plots of this section we use the average sysRatio which
is computed by taking for each value of $1\leq k\leq10$ the average
over the $N_{k,m}100=1000$ runs. Figure \ref{fig:Algorithmic-complexity-}
shows the graph (with $\mathsf{x}$) of the average estimated algorithmic
complexity $\ell_{0}$ of $\xi_{0}^{(n)}$ versus the average system
ratio $\rho$. The dashed lines are the upper and lower envelopes
of the estimated standard deviation from the mean. This variance arises
from the different values of training size $m$ and from the fact
that both the training and test sequences are random. The arrow points
at the value of $\rho^{*}$ that corresponds to $k^{*}=5$ (the source
model order). As can be seen, for low values of $\rho$ the spread
in $\ell_{0}$ is low. There is a critical point at $\rho^{*}$ where
the spread around the mean value of $\ell_{0}$ increases significantly
as $\rho$ increases. 

We know from section \ref{sec:Selection-rule} that the higher the
algorithmic complexity of a selection rule the higher the possible
deviation of the frequency of $1$s in the selected subsequence (the
stochastic deviation). As mentioned above, in \cite{rat0903} it was
shown that the decision rule of a learner can be represented as a
selection rule that picks the subsequence corresponding to the mistakes
made when predicting $0$s in the input test sequence. The theory
predicts that the stochastic deviation of the mistake sequence $\xi_{0}^{(n)}$
grows as the complexity of the decision rule increases. We now validate
this experimentally. 

\begin{figure}
\subfloat[Gzip as compressor]{\includegraphics[clip,scale=0.6]{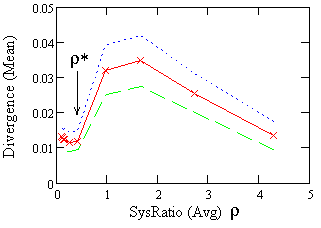}

}\subfloat[PPM as compressor]{\includegraphics[clip,scale=0.6]{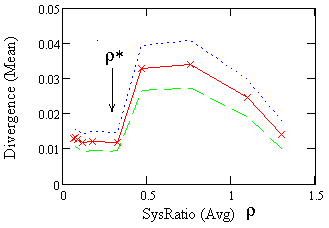}

}

\caption{\label{fig:Divergence--of}Divergence $\Delta_{0}$ of the mistake
subsequence $\xi_{0}^{(n)}$ versus the sysRatio (averaged) }

\end{figure}

Figure \ref{fig:Divergence--of} displays the graph (with $\mathsf{x}$)
of the average divergence $\Delta_{0}$ of the mistake subsequence
$\xi_{0}^{(n)}$ versus the average of the sysRatio $\rho$ where
again averages are taken over the $1000$ runs as described above.
The dashed lines are the upper and lower envelopes of the standard
deviation from the mean. The arrow points at the value of $\rho^{*}$
that corresponds to $k^{*}$ (the source model order). As can be seen,
for low values of $\rho$ the spread of $\Delta_{0}$ is low. Similar
to the previous result for $\ell_{0}$, also here we see a relative
minimum at $\rho^{*}$ where the standard deviation around the mean
value of $\Delta_{0}$ increases once we increase $\rho$ beyond $\rho^{*}$.
Since we know there is an inverse relationship between $\rho$ and
$k$ (Figure \ref{fig:SysRatio--versus K}) then the small hook shape
that appears to the left of the plot in Figure \ref{fig:Divergence--of}
indicates an increase in the $\Delta_{0}$ value as $k$ increases
beyond $k^{*}$ ($\rho$ decreases below $\rho^{*}$). Thus data overfitting
(which occurs when $k>k^{*)}$ is depicted here via this slight increase
in the divergence $\Delta_{0}$ as we decrease $\rho$ beyond the
$\rho^{*}$.

It follows from this result that the sysRatio $\rho$ (which is a
measure of information density of the learner's model \cite{Ratsaby_infodensity09})
influences how random are the mistakes made by a learner. The sysRatio
$\rho$ is a proper measure of complexity of a learner's decision
rule since it is with respect to $\rho$ that the characteristics
of the random mistake subsequence $\xi_{0}^{(n)}$ are consistent
with the theory \cite{rat0903,ratsabyICCC09}, namely, the higher
the sysRatio the more significant the deviation $\Delta_{0}$ of $\xi_{0}^{(n)}$
from a pure Bernoulli random sequence. 

We have so far considered $\rho$ as an independent variable. In section
\ref{sub:The-sysRatio-surface} we study the sysRatio as a dependent
variable, i.e., as a function of the estimated algorithmic complexity
$\ell_{0}$ and divergence $\Delta_{0}$. Before looking at that we
proceed to show how $\ell_{0}$ varies with respect to the error $p_{0}$
which will now play the role of the independent variable.

\subsection{\label{sub:Estimated-algorithmic-complexity_DIF}Estimated algorithmic
complexity $\ell_{0}$ versus the error $p_{0}$ for different values
of $k$ }

\begin{figure}
\begin{centering}
\includegraphics[clip,scale=0.6]{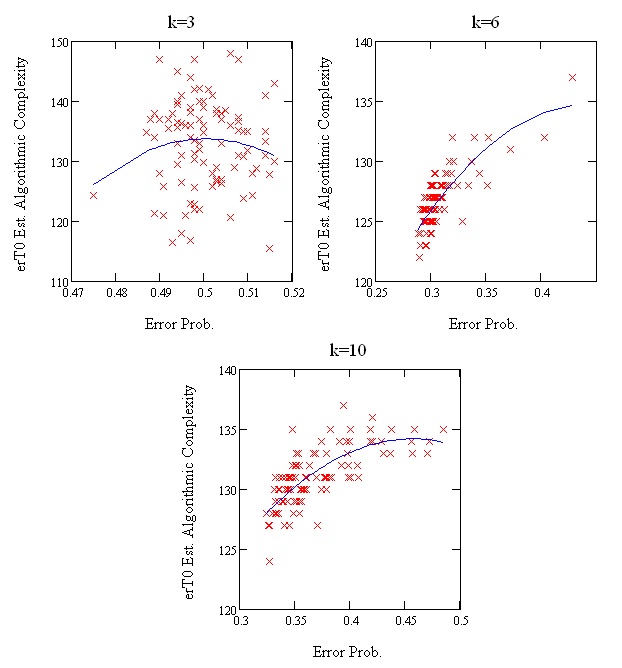}
\par\end{centering}

\caption{\label{fig:Error-versus-estimated}Estimated algorithmic complexity
$\ell_{0}$ versus error probability, for $k=3$, $6$ and $10$ and
$k^{*}=5$ , Gzip-based  compressor}

\end{figure}

We first mention that in all the figures below we reduced the number
of data points (using simple random sampling) for clarity of presentation.
Figure \ref{fig:Error-versus-estimated} shows the estimated algorithmic
complexity $\ell_{0}$ of the mistake subsequence $\xi_{0}^{(n)}$
versus the probability of error $p_{0}$. The curves are a second
order regression. For $k=3<k^{*}$ there is no clear relationship
but for $k=6$ (just above $k^{*})$ we see a sharp rise in $\ell_{0}$
with respect to an increasing $p_{0}$ (the regression polynomial
is: $-448x^{2}+396x+47$). When $k=10$ (double the value of $k^{*})$
we see a less steep increase (the regression polynomial is: $-356x^{2}+325x+60$). 

\subsection{Estimated algorithmic complexity $\ell_{0}$ and divergence $\Delta_{0}$
versus error $p_{0}$ over full range of $m$ and $k$ }

\begin{figure}
\begin{raggedright}
\subfloat[$\ell_{0}$ (marked $\mathsf{x}$) compared to entropy-based estimate
(marked $+$) versus $p_{0}$]{\includegraphics[clip,scale=0.5]{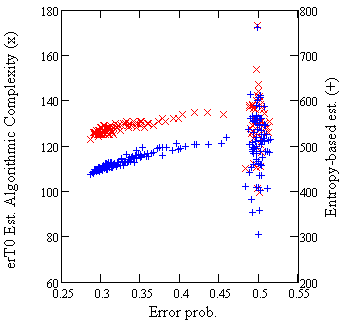}

}\subfloat[$\ell_{0}$ (marked $\mathsf{x})$ compared to $\Delta_{0}$ (marked
$\circ$) versus $p_{0}$]{\includegraphics[clip,scale=0.5]{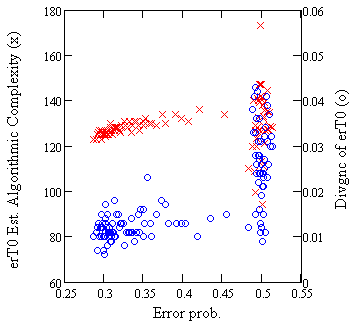}

}
\par\end{raggedright}

\caption{\label{fig:(a)-Estimated-algorithmic}Estimated algorithmic complexity
$\ell_{0}$ and divergence $\Delta_{0}$ as function of $p_{0}$ ,
Gzip-based compressor, for population of error sequences based on
learners with $m=100,200,\ldots,10,000$ and models of order $k=1,2,\dots10$. }

\end{figure}

In Figure \ref{fig:(a)-Estimated-algorithmic}(A) we compare $\ell_{0}$
marked in red ($\mathsf{x}$) to the entropy-based estimate of (\ref{eq:entropyest})
marked in blue ($+$) where we substitute for $n$ in (\ref{eq:entropyest})
the length $n_{0}$ of the sequence $\xi_{0}^{(n)}$ and the probability
$p_{0}$ for the parameter $p$. The value of the Pearson's correlation
coefficient between $\ell_{0}$ and the entropy estimate is $0.925$
indicating a high correlation (almost linear). Thus the entropy-based
estimate appears to be good for the whole population of learners which
consists of training sequences of size $m=100,200,\ldots,10,000$
and models of order $k=1,2,\dots10$. In Figure \ref{fig:(a)-Estimated-algorithmic}(A)
for the $\ell_{0}$ data (marked by $\mathsf{x})$ there appear to
be two clusters of points (sequences) separated by an error probability
gap at $p_{0}\approx0.45$. The first region is for $p_{0}<0.38$.
We refer to it as the \emph{cool} cluster. Here the complexity $\ell_{0}$
values are concentrated. The other cluster (termed \emph{hot}) is
where $p_{0}>0.45$. Here the spread in values of $\ell_{0}$ is significantly
larger than in the cool cluster.

In Figure \ref{fig:(a)-Estimated-algorithmic}(B) we see that the
divergence $\Delta_{0}$ (marked by the symbols $\circ$) and the
complexity $\ell_{0}$ (marked by $\mathsf{x}$) are somewhat correlated
(Pearson's coefficient of $0.241$) and it is due to the fact that
the divergence values $\Delta_{0}$ are also split into two clusters
which are in correspondence with the two clusters of the $\ell_{0}$
values. 

\begin{figure}
\subfloat[Gzip as compressor]{\includegraphics[clip,scale=0.6]{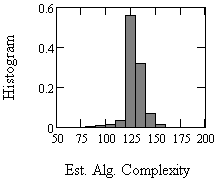}

}\subfloat[PPM as compressor]{\includegraphics[clip,scale=0.6]{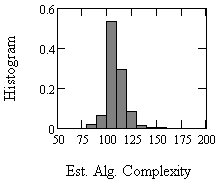}

}

\caption{\label{fig:Histogram-of-}Histogram of $\ell_{0}$ for population
of error sequences based on learners with $m=100,200,\ldots,10,000$
and models of order $k=1,2,\dots10$. }

\end{figure}

Let us look at the distribution of $\ell_{0}$ which is shown in Figure
\ref{fig:Histogram-of-}. The distribution is very similar for both
types of compressors. For the Gzip-based and PPM-based compressors
the mean values are $\mu_{0}=129$, $109$ and the distributions have
skewness of $1.79$, $1.8$ and kurtosis of $2.39$, $2.66$, respectively
( for the normal distribution the skewness and kurtosis are $0)$.
This indicates that the distributions are positively asymmetric (a
heavier right tail) and peaked.

\subsection{\label{sub:The-error-surface}The error $p_{0}$ surface}

\begin{figure}
\subfloat[Gzip based compressor]{\includegraphics[clip,scale=0.6]{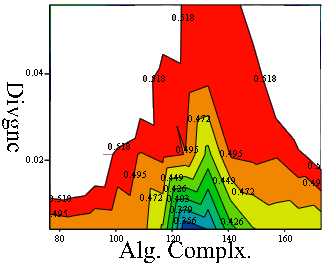}

}\subfloat[PPM based compressor]{\includegraphics[clip,scale=0.6]{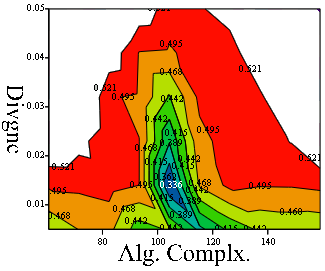}

}

\caption{\label{fig:3DplotofErrorVsDivgncAlgComp}Error probability $p_{0}$
as a function of divergence $\Delta_{0}$ and estimated algorithmic
complexity $\ell_{0}$}

\end{figure}

Figure \ref{fig:3DplotofErrorVsDivgncAlgComp} depicts the first central
result of the paper. It displays the error probability $p_{0}$ as
a function of the divergence $\Delta_{0}$ and estimated algorithmic
complexity $\ell_{0}$ (we note that the jagged contour lines are
due to the interpolation mesh being limited in size and do not reflect
actual data). At the center bottom we see the contour level of $0.356$
(this is approximately the Bayes error level) and the topmost contour
is at a value of $0.518$ which corresponds to prediction by pure-guessing.
We can ascertain the following from this interesting plot: the population
of mistake sequences of lowest error probability (close to the Bayes
$0.3$ value) concentrates close to the mean value $\mu_{0}$ and
has a very low divergence $\Delta_{0}$. This region corresponds to
the cool cluster of Figure \ref{fig:(a)-Estimated-algorithmic} (we
call it the cool region and it appears in blue in Figure \ref{fig:3DplotofErrorVsDivgncAlgComp}).
This characteristic indicates that the sequences in the cool region
are close to being truly random Bernoulli sequences with parameter
$p_{0}$. As we start to look at a population of sequences with a
higher error probability $p_{0}$ and walk along its fixed contour
level we have a tradeoff between two possible choices: (1) to have
a complexity $\ell_{0}$ value which is far from the mean (less than
or greater than $\mu_{0}$) and maintain a low divergence $\Delta_{0}$
value or (2) to have a large divergence $\Delta_{0}$ and maintain
an $\ell_{0}$ which is close to the mean $\mu_{0}$. The union of
the red and orange regions in Figure \ref{fig:3DplotofErrorVsDivgncAlgComp}
corresponds to the hot cluster that we saw in Figure \ref{fig:(a)-Estimated-algorithmic}.
By definition of the maximum \emph{a posteriori }probability decision
rule that we are using the error can never exceed $0.5$ so the true
error surface cannot exceed $0.5$ and this is why we see that the
empirical error surface ends at a contour level close to $0.5$. 

An interesting point that we see here is that this surface is defined
only over a part (colored region) of the $\Delta\ell$-plane. We term
this the \emph{admissible} region of the $\Delta\ell$ plane and it
is induced by the error surface. In Figure \ref{fig:3DplotofErrorVsDivgncAlgComp}
we see that the contour area is slightly larger on the right side
of $\mu_{0}$ than on the left of $\mu_{0}$ which is consistent with
the heavier right tail of the $\ell_{0}$ distribution in Figure \ref{fig:Histogram-of-}.
So admissibility appears to have a slight intrinsic bias towards complexity
values $\ell_{0}$ that are larger than the mean $\mu_{0}$.

If we regard sequences in the the cool region as \emph{truly} \emph{random}
(i.e., having a complexity value $\ell_{0}$ close to the mean $\mu_{0}$
and a low divergence from Bernoulli) then we can introduce a new perspective
on the process of learning. When the process is perfect, it produces
a Bayes optimal predictor whose mistake sequence falls in the cool
region. But when it is imperfect (due to limited training size $m$
or improper model order $k$) the process produces a \emph{malformed}
sequence which is either atypically chaotic ($\ell_{0}$ far from
$\mu_{0}$) but stochastic (low $\Delta_{0}$) or typically chaotic
($\ell_{0}$ close $\mu_{0}$) but atypically stochastic (large $\Delta_{0}$).

So far we discussed the error surface which is intrinsically a property
of the random mistake sequence since $p_{0}$ is defined only based
on the ratio of the number of $1$s to the length of the sequence.
In the next section we examine the sysRatio surface which intrinsically
is a learner's characteristic since it measures the information density
$\rho$ of the learner's decision rule.

\subsection{\label{sub:The-sysRatio-surface}The sysRatio $\rho$ surface}

\begin{figure}
\subfloat[Gzip based compressor]{\includegraphics[clip,scale=0.65]{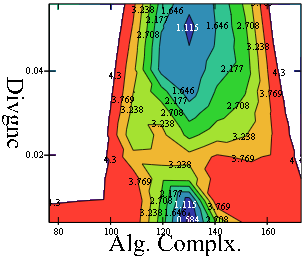}

}\subfloat[PPM based compressor]{\includegraphics[clip,scale=0.65]{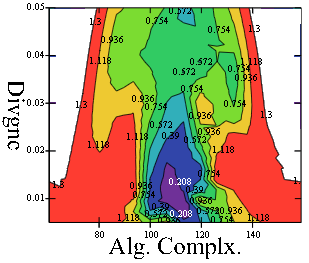}

}

\caption{\label{fig:Sysratio--asfunc_DVI_ALG}sysratio $\rho$ as a function
of divergence $\Delta_{0}$ and estimated algorithmic complexity $\ell_{0}$}

\end{figure}

Figure \ref{fig:Sysratio--asfunc_DVI_ALG} displays the next central
result of the paper, a contour plot of the sysRatio $\rho$ over the
$\Delta\ell$-plane. The outer contours (red) are for higher values
of $\rho$. There are two relative minima one of which is at a lower
value of $\Delta_{0}$ and touches the $\Delta_{0}=0$ axis while
the other appears above the $0.04$ divergence level. Based on what
we already know about $\rho$ versus $k$ (Figure \ref{fig:SysRatio--versus K})
we can conclude that the lower minimum in Figure \ref{fig:Sysratio--asfunc_DVI_ALG}
is in a region of the plane that corresponds to sequences generated
by learners of order $k$ which is equal to $k^{*}$ or just slightly
above $k^{*}$ (we call this region \emph{$OM$} for 'overfitting
minimum') while the upper relative minimum in Figure \ref{fig:Sysratio--asfunc_DVI_ALG}
is in the region of sequences generated by learners of order $k$
which is slightly lower than $k^{*}$ (we call this region $UM$ for
'underfitting minimum'). The remaining regions (colored green to red)
are where the learners have an order $k$ significantly less than
$k^{*}$. Thus there is a saddle point as one passes from $UM$ to
$OM$ and cross from $k$ which is just under $k^{*}$ to $k=k^{*}$.
This is more pronounced in the Gzip-based compressor than in the PPM-based
compressor.

Based on this plot we can see that a decision rule with a high information
density (sysRatio value $\rho$) yields an atypically chaotic random
error sequence, i.e., with an estimated algorithmic complexity value
$\ell_{0}$ that is far from the mean $\mu_{0}$. As the information
density of the decision rule decreases the complexity of the error
sequence moves towards a typical value ($\ell_{0}$ closer to $\mu_{0}$)
and its divergence from Bernoulli decreases towards zero. 

Recall from the end of section \ref{sec1} that the act of predicting
bits of the input test sequence $x^{n}$ to be $0$s is equivalent
to selecting from $x^{n}$ a subsequence $\xi_{0}^{(n)}$. We are
now in a position to understand that this selection process produces
random binary sequences $\xi_{0}^{(n)}$ of different character and
'spreads' them in different regions of the $\Delta\ell$-plane. This
spreading is a consequence of what we term \emph{scattering} bits
of a sequence since it resembles particle scattering in physics (it
is also similar to the concept of chaotic scattering \cite{chaoticscattering}
where instead of initial conditions of the learner we characterize
it by its information density $\rho$). Given a random input sequence
$x^{(n)}$ the learner (in his decision/selection action) effectively
scatters the bits of $x^{(n)}$ in a way that resembles the binary
collisions of particles in a beam with other particles that knock
the beam particles into different directions. From this scattering
the resulting sequence of bits is $\xi_{0}^{(n)}$. The learner here
acts as a static structure (a solid of some kind), or a localized
target such as a thin foil in a physical scattering experiment. Learners
with high information density$\rho$ scatter bits of the input sequence
more wildly thereby producing sequences (points in the $\Delta\ell$-plane)
that deviate from typical complexity values or have high stochastic
divergence. As mentioned in section \ref{sec:Overview} this is in
line with the model introduced in \cite{Ratsaby_entropy} where a
static structure is said to deform the randomness characteristics
of an input sequence of excitations.

It is interesting to ask at this point whether as a consequence of
this phenomenon it may perhaps be possible to optimally fine-tune
a learner's model-order $k$ just by observing the randomness characteristics
of the mistake error $\xi_{0}^{(n)}$, i.e., adjusting $k$ in a direction
that corresponds to decreasing $\rho$ towards the $OM$ region. It
is not yet clear whether such a scheme that monitors the random characteristics
of the mistake sequence would yield better performance (either accuracy
or computational efficiency) compared to doing standard model-selection
which adjusts $k$ directly based on some form of estimate of the
generalization error \cite{DevroyeGyorfiLugosi1996}.

\section{Conclusions}

This paper is an experimental investigation of the problem that was
posed and theoretically solved in \cite{rat0903,ratsabyICCC09}. We
have reconfirmed that the sysRatio $\rho$ originally introduced in
\cite{Ratsaby_Acta,Ratsabyuics09,Ratsaby_BRAIN10} is a proper measure
of the complexity of a learner's decision rule as it is with respect
to $\rho$ that the deformation of randomness of the mistake subsequence
$\xi_{0}^{(n)}$ takes place in consistence with the theory, namely,
the higher the value of $\rho$ the more significant the divergence
$\Delta_{0}$ of the mistake sequence $\xi_{0}^{(n)}$ relative to
a pure Bernoulli random sequence. The two central results introduced
in the current paper depict the special structure of the error probability
$p_{0}$ and sysRatio $\rho$ surfaces over the $\Delta\ell$-plane.
They imply that bad learners generate atypically complex or stochastically
divergent mistake sequences while good learners generate typically
complex sequences with low divergence from Bernoulli. Since a learner
can be modeled as a selection rule we name this phenomenon 'bit-scattering'.
The idea follows the general model of static algorithmic interference
introduced in \cite{Ratsaby_entropy} whereby effectively the learner
acts as a static structure whose complexity is the sysRatio (information
density $\rho$). It produces randomly-deformed types of mistake sequences
where deformation is proportional to $\rho$.

\appendix

\section{\label{sec:apA}$\;$}

\begin{figure}[H]
\begin{raggedright}
\subfloat[PPM-based compressor, $k^{*}=3$]{\noindent \begin{raggedright}
\includegraphics[clip,scale=0.7]{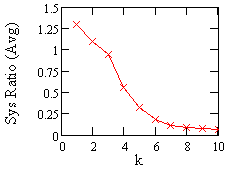}
\par\end{raggedright}

}\subfloat[PPM-based compressor, $k^{*}=7$]{\includegraphics[clip,scale=0.7]{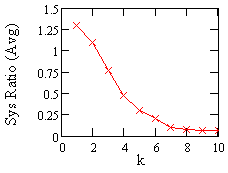}

}
\par\end{raggedright}

\caption{\label{fig:The-critical-point}Rates of change of $\rho$ with respect
to $k$}

\end{figure}

In this section we present some additional auxiliary results pertaining
to the relationship between the sysRatio $\rho$ and model order $k$.
In section \ref{sub:sysRatio--versus} for a learning problem with
$k^{*}=5$ we saw that for the PPM-based compressor the graph of the
average $\rho$ versus $k$ is decreasing and has a critical point
in the vicinity of $k^{*}$. Figure \ref{fig:The-critical-point}
shows that this critical point also appears in learning problems with
$k^{*}=3$. For $k^{*}=7$ there appears to be two critical points,
one of which is at $k^{*}$.

\bibliographystyle{plain}

\end{document}